\documentclass{article}
\usepackage{ijcai21}

\usepackage{times}
\usepackage{soul}
\usepackage{url}
\usepackage[hidelinks]{hyperref}
\usepackage[utf8]{inputenc}
\usepackage[small]{caption}
\usepackage{graphicx}
\graphicspath{ {./images/} }
\usepackage{amsmath}
\usepackage{amsthm}
\usepackage{booktabs}
\usepackage{algorithm}
\usepackage{algorithmic}
\usepackage{multirow}
\usepackage{amssymb}
\usepackage{makecell}
\usepackage{bm}
\usepackage{color}

\makeatletter 
  \newcommand\figcaption{\def\@captype{figure}\caption} 
  \newcommand\tabcaption{\def\@captype{table}\caption} 
\makeatother

\title{Few-Shot Learning with Part Discovery and Augmentation \\ from Unlabeled Images}

\author{
Wentao Chen$^{1,2}$\and
Chenyang Si$^{2}$\and
Wei Wang$^{2}$\and
Liang Wang$^{2}$\and
Zilei Wang$^1$\and
Tieniu Tan$^{1,2}$\footnote{Contact Author}
\affiliations
$^1$University of Science and Technology of China\\
$^2$Center for Research on Intelligent Perception and Computing, NLPR, CASIA\\
\emails
\{wentao.chen, chenyang.si\}@cripac.ia.ac.cn,
\{wangwei, wangliang, tnt\}@nlpr.ia.ac.cn,
zlwang@ustc.edu.cn
}

\begin{document}

\maketitle

\begin{abstract}
  
  
  Few-shot learning is a challenging task since only few instances are given for recognizing an unseen class. One way to alleviate this problem is to acquire a strong inductive bias via meta-learning on similar tasks. In this paper, we show that such inductive bias can be learned from a flat collection of unlabeled images, and instantiated as transferable representations among seen and unseen classes.   Specifically, we propose a novel part-based self-supervised representation learning scheme to learn transferable representations by maximizing the similarity of an image to its discriminative part. 
  To mitigate the overfitting in few-shot classification caused by data scarcity, we further propose a part augmentation strategy by retrieving extra images from a base dataset.
  We conduct systematic studies on \emph{mini}ImageNet and \emph{tiered}ImageNet benchmarks. Remarkably, our method yields impressive results, outperforming the previous best unsupervised methods by 7.74\% and 9.24\% under 5-way 1-shot and 5-way 5-shot settings, which are comparable with state-of-the-art supervised methods.
  
  
  
\end{abstract}

\section{Introduction}


Recently, great progress in the computer vision community has been achieved with deep learning, which often needs numerous training data. Unfortunately, there are many practical applications where collecting data is very difficult. To learn a novel concept with only few examples, few-shot learning has been recently proposed and gains extensive attention.\cite{Doersch2020CrossTransformersSF,Afrasiyabi2020AssociativeAF}


A possible solution to few-shot learning is meta-learning \cite{pmlr-v70-finn17a,vinyals2016matching,Ravi2017OptimizationAA,snell2017prototypical,peng2019few}. It first extracts shared prior knowledge from many similar tasks. After that, this knowledge will be adapted to the target few-shot learning task to restrict the hypothesis space, which makes it possible to learn a novel concept with few examples. Equipped with neural networks, the prior knowledge can be parameterized as an embedding function, or a set of initial parameters of a network, which often needs an extra labeled base dataset for training. 
Another line of work is based on transfer learning \cite{10.1007/978-3-030-58568-6_16,Dhillon2020A}, which extracts prior knowledge as a pre-trained feature extractor. Typically, a standard cross entropy loss is employed to pre-train the feature extractor, which also needs a labeled base dataset. However, collecting a large labeled dataset is time-consuming and laborious. Besides, these labels are sadly discarded when performing a target few-shot learning task, because they belong to different class spaces. Inspired by recent progress of unsupervised learning, a question is naturally asked: can we can learn prior knowledge only from unlabeled images? If yes, it will be a promising approach for the scenario where many unlabeled images are available but the target task is data-scarce.

Some remarkable works have made an effort for this purpose, e.g., unsupervised meta-learning \cite{hsu2018unsupervised,khodadadeh2019unsupervised}. However, these unsupervised methods are hindered by learning effective class-related representations from images, compared to the supervised counterparts. This is because much unrelated or interfering information, e.g., background clutters, may 
impose adverse impacts on representation learning
under label-free unsupervised setting. Selecting discriminative image regions or target parts is an effective way to reduce this interference during representation learning, which has a consistent motivation with traditional part-based models \cite{pedro2009part}.


In this paper, we propose a part-based self-supervised learning model, namely Part Discovery Network (PDN), to learn more effective representations from unlabeled images. The key point of this model is mining discriminative target parts from images. Due to the lack of part labels, multiple image regions are first extracted via random crop, which inevitably contain interfering backgrounds or less informative regions. To eliminate these regions, we choose the image region as the most discriminative target part, which keeps the largest average distance to other images (negative samples). The rationale of this selection is that the discriminative part should be able to distinguish the original image from others \cite{singh2012unsupervised}, so its average distance to other images should be large enough. With the selected discriminative part, we maximize its similarity to the original image in a similar way to \cite{He2020MomentumCF}.

\begin{figure}
    \centering
    \includegraphics[width=8cm]{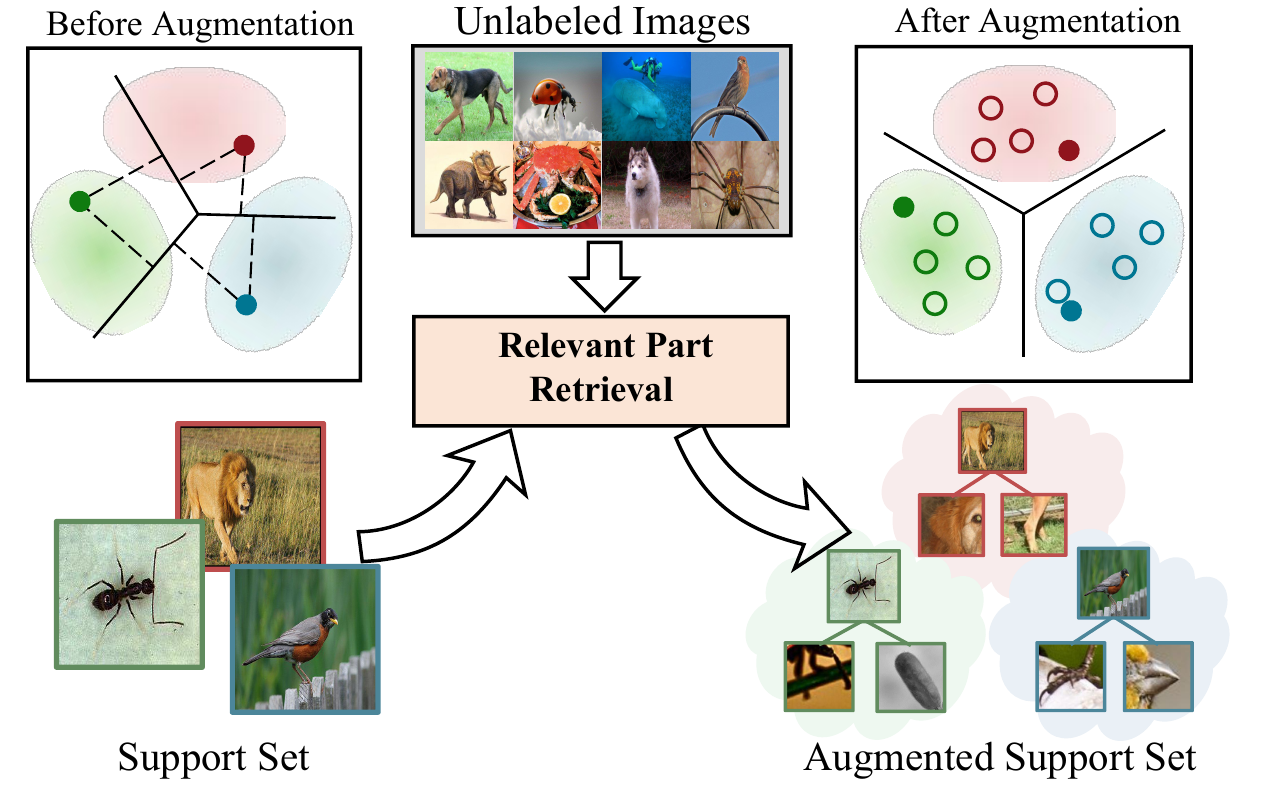}
    \caption{The main idea of the proposed Part Augmentation Network. The classifier is easy to overfit due to the scarcity of support samples in few-shot scenario. We retrieve relevant parts from unlabeled images to augment support set. The augmented support set can help learn a robust classifier with clear decision boundary.} 
    \label{fig:idea}
\end{figure}

Another challenge in the few-shot scenario is that the target classifier is easy to overfit due to the scarcity of support training samples. An effective way to prevent overfitting is data augmentation, which has been explored in the few-shot learning literature \cite{NEURIPS2018_1714726c,wang2018low}. However, most of these methods assume that the variance of seen classes can be directly transferred to unseen classes, which is too strict in most situations. 
In this paper, we resort to retrieve extra images from a base dataset and extract their part features to augment support set, based on the fact that similar objects generally share some common parts. 
The core of our method is selecting these part features which match well with image features in the support set.
Specifically, we propose a novel Class-Competitive Attention Map ($\rm C^2AM$) to guide the relevant part selection from the retrieved images, and then refine target classifier with these selected parts. 
Our method is also called Part Augmentation Network (PAN), and Figure \ref{fig:idea} illustrates the main idea of PAN.


Our Part Discovery and Augmentation Network (PDA-Net) consisting of both PDN and PAN largely outperforms the stat-of-the-art unsupervised few-shot learning methods, and achieves the comparable results with most of the supervised methods. The contributions of this work can be summarized as:


\begin{itemize}
    \item We propose a novel self-supervised Part Discovery Network, which can learn more effective representations from unlabeled images for few-shot learning.
    \item We propose a Part Augmentation Network to augment few support examples with relevant part features, which mitigates overfitting and leads to more accurate classification boundaries.
    \item Our method outperforms previous unsupervised methods on two standard few-shot learning benchmarks. Remarkably, our unsupervised PDA-Net is also comparable with supervised methods.
\end{itemize}



\section{Related Work}
\paragraph{Few-Shot Learning.} Few-shot learning aims at learning a novel concept from few examples. A possible solution is meta-learning, which extracts prior knowledge from many similar tasks (episodes). For example, MAML \cite{pmlr-v70-finn17a} learns the optimal initial parameters of a network, which can be quickly adapted to a new task via gradient descent. Matching Networks \cite{vinyals2016matching} classifies an instance based on its nearest labeled neighbor in the learned embedding space. Different from meta-learning, \cite{Dhillon2020A} demonstrate that a strong transfer learning baseline can achieve competitive performance.
In order to train models, all of the above methods need a large labeled base dataset. Unsupervised  meta-learning  methods loose  this constraint by constructing training episodes from unlabeled images. \cite{hsu2018unsupervised} propose to acquire pseudo labels by clustering in an unsupervised feature space. \cite{khodadadeh2019unsupervised} use random sampling and augmentation to create synthetic episodes. Our method aims at both learning more effective representation from unlabeled images and increasing support set by data augmentation for unsupervised few-shot learning.

\begin{figure*}[t]
    \centering
    \includegraphics[width=15cm]{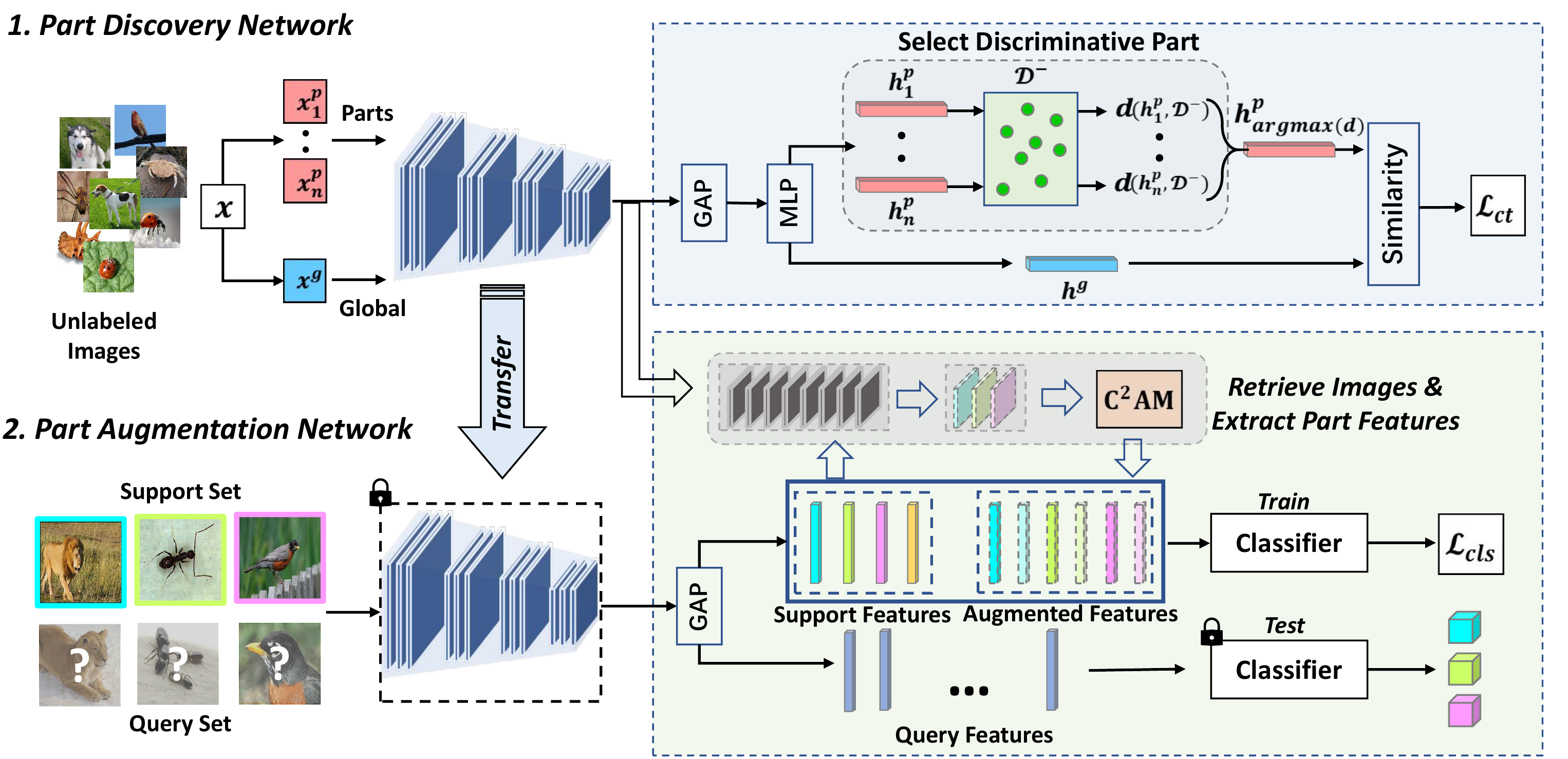}
    \caption{The framework of the proposed Part Discovery and Augmentation Network. In Part Discovery Network (PDN), we first extract multiple parts from an image, and then select the most discriminative one. Through maximizing the similarity of the global view of the image to the selected discriminative part, we learn effective representations from unlabeled images. In Part Augmentation Network (PAN), we retrieve relevant images to the support set from unlabeled images, and then create a Class-Competitive Attention Map ($\rm C^2AM$) to select relevant parts as augmented features. Finally, a classifier is trained on both support features and augmented features. Note that the feature extractor in PAN is transferred from PDN.}
    \label{fig:framework}
\end{figure*}



\paragraph{Self-Supervised Learning.} Self-supervised learning aims at leveraging unlabeled images to learn good representations for down-stream tasks. Previous work mainly focuses on mining supervision signals from unlabeled data \cite{gidaris2018unsupervised,Doersch2015UnsupervisedVR,Zhang2016ColorfulIC}. Recently, contrastive learning shows superior performance improvement, which maximizes the similarity of two different views of the same image on global level \cite{Chen2020ASF} or local level \cite{ouali2020spatial}, or enforces consistency of cluster assignments between two views \cite{Caron2020UnsupervisedLO}. In this paper, we explicitly mine discriminative parts for contrastive learning, which learns more effective representations from unlabeled images.

\paragraph{Data Augmentation for Few-Shot Learning.} A straightforward way to deal with data deficiency is to synthesize more data. \cite{NEURIPS2018_1714726c} propose to extract intra-class variance (deltas) from base classes and use them to synthesize samples for novel classes. \cite{wang2018low} encode the intra-class variance in a hallucinator, which can synthesize new features taking as input a reference feature and noise vectors. To leverage base class samples, \cite{Chen2019ImageDM} propose to generate a deformed image by fusing a probe image and a gallery image. 
Our method augments support set by retrieving extra images from a base dataset and extracting matched part features with the support samples.


\section{Notation}
In this section, we briefly illustrate the formulation of the few-shot image classification problem. Given a labeled support set $\mathcal{S}=\{(\bm{x_i}, y_i)\}_{i=1}^{N_s}$ where $\bm{x_i}\in I$ is an image and $y_i\in \mathcal{C}_{novel}$ is its label, we are supposed to predict the labels of a query set $\mathcal{Q}=\{(\bm{x_i}, y_i)\}_{i=1}^{N_q}$, which also belongs to $ \mathcal{C}_{novel}$. The number of classes $|\mathcal{C}_{novel}|$ is called \emph{way} and the number of samples in each class is called \emph{shot}. For few-shot learning, the shot is very small, like 1-shot or 5-shot. Due to the scarcity of support samples, it is very hard to train a classification model from scratch. Therefore, we are given an extra large base dataset $\mathcal{D}^{base}$ to mine prior knowledge. Here, $\mathcal{D}^{base}$ is from base classes $\mathcal{C}_{base}$, and $\mathcal{C}_{base}\cap \mathcal{C}_{novel}=\phi$. Previous works usually need base labels to construct training episodes or pre-train the classification model. In this paper, we only use unlabeled images in $\mathcal{D}^{base}$, i.e., $\mathcal{D}^{base}=\{\bm{x_i}\}_{i=1}^{N_b}$. 


\section{Method}
Our method follows a transfer learning protocol which includes two stages, namely the representation learning stage and the few-shot learning stage. On the first stage, we aim at learning a feature extractor $f$ from $\mathcal{D}^{base}$.
On the second stage, the learned feature extractor $f$ is transferred to the target few-shot learning task, followed by training a linear classifier 
using support set $\mathcal{S}$. The final classification results of query set $\mathcal{Q}$ are predicted by the learned classifier. From the above description we can see that the key is to learn a good feature extractor from base dataset $\mathcal{D}^{base}$ and meanwhile obtain a robust classifier with limited data in support set $\mathcal{S}$. To this end, we propose Part Discovery and Augmentation Network (PDA-Net), which consists of a Part Discovery Network (PDN) to learn effective representations, and a Part Augmentation Network (PAN) to augment few support examples with relevant part features. The whole framework is shown in Figure \ref{fig:framework}.


\subsection{Part Discovery Network}\label{pdn}

Our Part Discovery Network (PDN) is a part-based self-supervised learning model, which maximizes the similarity between representations of the global view of an image and its discriminative part. We will introduce the details of selecting discriminant part as follows.


\paragraph{Extracting Multiple Parts.} Without part labels, we generate parts by randomly cropping a given image $\bm{x}$ into $n$ patches \{$\bm{x^p_i}\}^n_{i=1}$. Meanwhile, a larger crop $\bm{x^g}$ is obtained to serve as global context. Random transformations are further applied to increase data diversity. To get latent part representations $\{\bm{h^p_i}\}^n_{i=1}$ and global representation $\bm{h^g}$, a convolution neural network $f$ is applied followed by global average pooling and a MLP projection head.

\paragraph{Selecting Discriminative Part.} Since we generate multiple parts with random crop, there are inevitably some crops that belong to background. Directly matching these crops to the global view will create bias towards backgrounds and hurt the generalization of the learned representations. To solve this problem, we develop an effective strategy to select the most discriminative part. Given a set of negative samples $\mathcal{D^-}=\{\bm{h_i^{g-}}\}_{i=1}^{N^-}$ that are exclusive with the input image $\bm{x}$, we define a sample-set distance metric $d(\bm{h^p_i},\mathcal{D^-})$ in the feature space, which indicates the distance between a part and all negative samples. We select the part with the maximum distance as the most discriminative one, therefore excluding background crops and less informative ones. The strategy is formulated as
\begin{equation}
    \bm{h^p} := \bm{h^p_{i^*}}, \quad where\ i^*=\arg\max_i d(\bm{h^p_i},\mathcal{D^-})
\end{equation}
where $\bm{h^p}$ is the selected part. The rationale of this strategy is that the discriminative part should be able to distinguish the original image from others, so its distance to other images should be very large.

The distance between $\bm{h^p_i}$ and $\mathcal{D^-}$ can be defined in many forms. One could use the minimum distance between $\bm{h^p_i}$ and all samples in $\mathcal{D^-}$. However, some similar images may exist in $\mathcal{D^-}$, so the minimum distance is severely affected by these similar images and can not reflect the true distance to most negative samples. 
Here we choose the mean distance to calculate $d(\bm{h^p_i},\mathcal{D^-})$,  which considers more on the samples of other classes.
$d(\bm{h^p_i},\mathcal{D^-})$ is calculated as:
\begin{equation}
    d(\bm{h^p_i},\mathcal{D^-}) = \frac{1}{|\mathcal{D^-}|}\sum_{\bm{h^{g-}}\in\mathcal{D^-}}-s(\bm{h^p_i}, \bm{h^{g-}})
\end{equation}
where $|\mathcal{D^-}|$ is the number of negative samples in this set, and $s$ is the cosine similarity.



\paragraph{Training.} 
With selected discriminative parts, we train the PDN with a contrastive loss, which is formulated as
\begin{equation}
    \resizebox{1\linewidth}{!}{$
    \mathcal{L}_{ct} = -\log\frac{\exp(s(\bm{h^p},\bm{h^g})/\tau)}{\exp(s(\bm{h^p,h^g})/\tau+\sum_{\bm{h^{g-}}\in\mathcal{D^-}}\exp(s(\bm{h^p},\bm{h^{g-}})/\tau)}
$}
\end{equation}
where $\tau$ denotes a temperature hyper-parameter.

To get a large negative set, we follow \cite{He2020MomentumCF} to organize $\mathcal{D^-}$ as a queue and use a momentum encoder to get consistent negative representations. The momentum encoder is an exponential moving average version of the feature extractor and MLP head. Its parameter $\theta_m$ is updated as
\begin{equation}
    \theta_m \longleftarrow m\theta_t + (1-m)\theta_m 
\end{equation}
where $m$ is a momentum hyper-parameter, and $\theta_t$ denotes the parameters of the feature extractor and MLP head at training step $t$.

\begin{figure}[t]
    \centering
    \includegraphics[width=7.5cm]{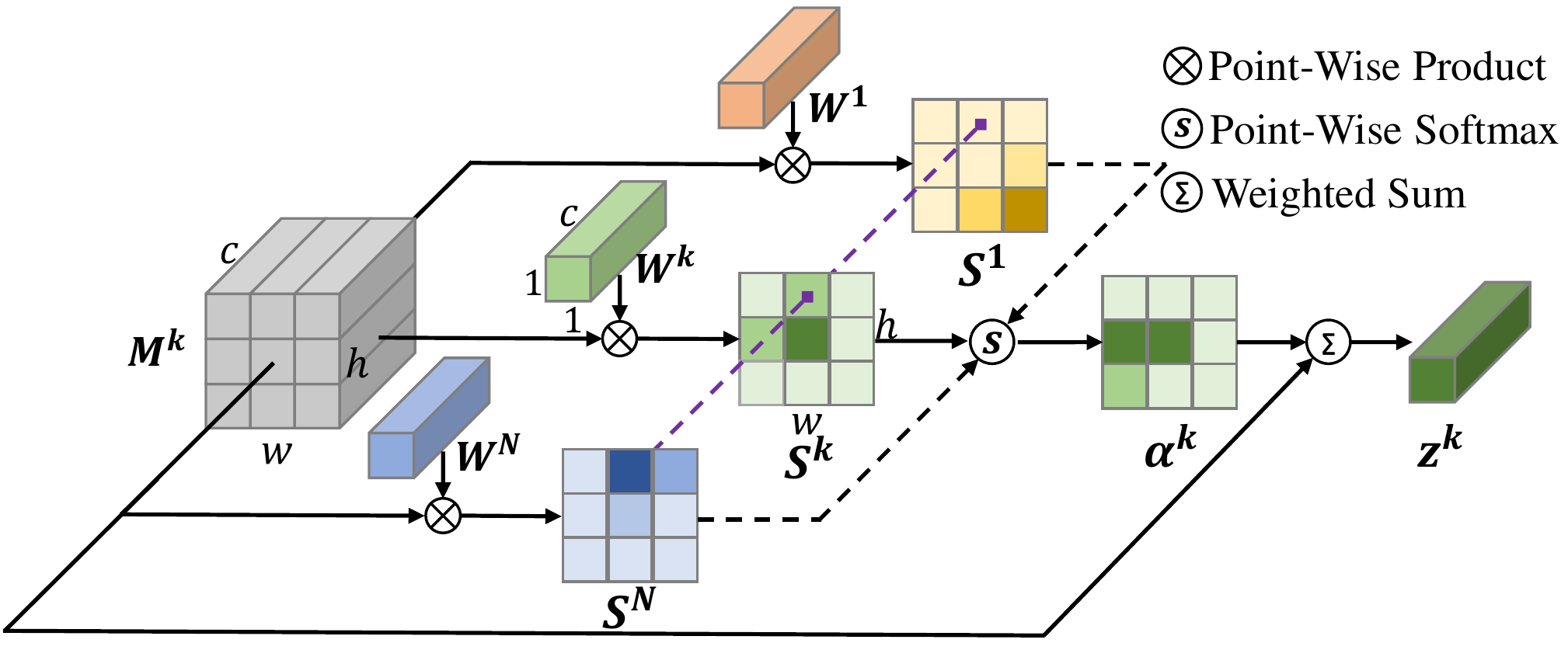}
    \caption{Illustration of $\rm C^2AM$. We omit bias item for simplicity.}
    \label{fig:CAM}
\end{figure}

\subsection{Part Augmentation Network} \label{pan}

Our Part Augmentation Network (PAN) aims to augment support set with relevant part features from base dataset. To this end, we first retrieve extra images from base dataset, then create a Class-Competitive Attention Map ($\rm C^2AM$) to guide the relevant part selection from the retrieved images, and finally refine target classifier with these selected parts.

\paragraph{Retrieving Extra Images.} Since the size of $\mathcal{D}^{base}$ is very large, we propose a simple but effective strategy to select a small number of very similar images from $\mathcal{D}^{base}$, which are more likely to contain relevant parts. Specifically, we first train a linear classifier $p(y|\bm{z}; \bm{W},\bm{b})$
on support set $\mathcal{S}$, where $\bm{z}$ denotes a feature vector, $\bm{W}$ and $\bm b$ are weight matrix and bias, respectively. Then, we employ this classifier to classify each image in $\mathcal{D}^{base}$ as:
\begin{equation}
    \hat{y}=\arg\max_i p(y=i|\bm{z}=GAP(\bm{M}))
\end{equation}
where GAP is a global average pooling operator and $\bm{M}$ denotes the feature map extracted by the learned feature extractor $f$ in Section \ref{pdn}

Among the images of class $k$ in $\mathcal{D}^{base}$, we keep the $N_a$ images with the highest classification probability as the retrieval results. The feature maps of these $N_a$ images are denoted as $\mathcal{A}_k = \{\bm{M_i^k}\}_{i=1}^{N_a}$.


\begin{table*}
\centering
\resizebox{1.\linewidth}{!}{
\begin{tabular}{lllcccc}
\toprule
 & & & \multicolumn{2}{c}{\emph{mini}ImageNet 5-way} &  \multicolumn{2}{c}{\emph{tiered}ImageNet 5-way} \\
Setting & Method  & Backbone & 1-shot & 5-shot & 1-shot & 5-shot \\
\midrule
\multirow{7}{*}{Supervised} 
 & IDeMe-Net \cite{Chen2019ImageDM} & ResNet-12 & 59.14$\pm$0.86 & 74.63$\pm$0.74 & - & - \\
 & MetaOptNet \cite{Lee2019MetaLearningWD} & ResNet-12 & 62.64$\pm$0.61 & 78.63$\pm$0.46 & 65.99$\pm$0.72 & 81.56$\pm$0.53 \\
 & Distill \cite{10.1007/978-3-030-58568-6_16} & ResNet-12 & 64.82$\pm$0.60 & 82.14$\pm$0.43 & 71.52$\pm$0.69 & 86.03$\pm$0.49 \\
 & Finetune \cite{Dhillon2020A} & WRN-28-10 & 57.73$\pm$0.62 & 78.17$\pm$0.49 & 66.58$\pm$0.70 & 85.55$\pm$0.48 \\
 & LEO \cite{rusu2018metalearning} & WRN-28-10 & 61.76$\pm$0.08 & 77.59$\pm$0.12 & 66.33$\pm$0.05 & 81.44$\pm$0.09 \\
 & CC+rot \cite{gidaris2019boosting} & WRN-28-10 & 62.93$\pm$0.45 & 79.87$\pm$0.33 & 70.53$\pm$0.51 & 84.98$\pm$0.36 \\
 & Align \cite{Afrasiyabi2020AssociativeAF} & WRN-28-10 &\textbf{65.92$\pm$0.60} & \textbf{82.85$\pm$0.55} & \textbf{74.40$\pm$0.68} & \textbf{86.61$\pm$0.59} \\
\midrule
\multirow{6}{*}{Unsupervised} & CACTUs \cite{hsu2018unsupervised} & Conv4 & 39.90$\pm$n/a & 53.97$\pm$n/a & - & - \\
 & UMTRA \cite{khodadadeh2019unsupervised} & Conv4 & 39.93$\pm$n/a & 50.73$\pm$n/a & - & - \\
 & Rot \cite{gidaris2019boosting} & WRN-28-10 & 43.43$\pm$n/a & 60.86$\pm$n/a & - & - \\
 & GdBT2 \cite{Khoi2021SSLGAN} & SN-GAN & 48.28$\pm$0.77 & 66.06$\pm$0.70 & 47.86$\pm$0.79 & 67.70$\pm$0.75 \\
 & MoCo \cite{10.1007/978-3-030-58568-6_16} & ResNet-50 & 54.19$\pm$0.93 & 73.04$\pm$0.61 & - & - \\
 & CMC \cite{10.1007/978-3-030-58568-6_16} & ResNet-50 & 56.10$\pm$0.89 & 73.87$\pm$0.65 & - & - \\
\cmidrule{2-7}
 & PDA-Net (Ours) & ResNet-50 & \textbf{63.84$\pm$0.91} & \textbf{83.11$\pm$0.56} & \textbf{69.01$\pm$0.93} & \textbf{84.20$\pm$0.69} \\
\bottomrule
\end{tabular}
}
\caption{Comparison with prior work on \emph{mini}ImageNet and \emph{tiered}ImageNet. Accuracy is reported with 95\% confidence intervals.}
\label{tab:comparison}
\end{table*}


\paragraph{Class-Competitive Attention Map} To further extract relevant parts from retrieved feature maps $\mathcal{A}_k$, we propose a novel CAM-based \cite{zhou2016learning} attention mechanism, Class-Competitive Attention Map ($\rm C^2AM$), illustrated in Figure \ref{fig:CAM}. Given a feature map $\bm{M^k}\in \mathcal{A}_k$, we obtain a class attention map $\bm{S^k}$ that indicates its relevance to class $k$ at each spatial location:
\begin{equation}
    \bm{S^k}(i,j) = \bm{W^k M^k}(i,j)+b^k
\end{equation}
where $\bm{S^k}(i,j)$ and $\bm M^k(i,j)$ are the classification score for class $k$ and the feature vector at location $(i,j)$, respectively. $\bm{W^k}, b^k$ are the classifier weight vector and bias for class $k$, which have been learned in $p(y|\bm{z}; \bm{W},\bm{b})$.

Although the class attention map $\bm{S^k}$ is very useful to locate class-specific image regions or object parts, it still contains some parts that have high classification scores for all classes. These parts provide less information for classification, and should be further inhibited. We perform softmax over the classification score vector $\bm{S}(i,j)$ at each spatial location, which provides a class competitive mechanism to highlight the parts which only have higher score for class $k$. The class-competitive attention map $\bm{\alpha^k}$ for class $k$ is as follows:
\begin{equation}
    \bm{\alpha^k}(i,j) = \frac{exp(\bm{S^k}(i,j))}{\sum_c^{|\mathcal{C}_{novel}|} exp(\bm{S^c}(i,j))}
\end{equation}


With this revised attention map $\bm{\alpha^k}$, we can extract more relevant part features $\bm{z^k}$ to augment the class $k$, which is calculated by the weighted sum of feature map $\bm{M^k}$:
\begin{equation}
    \bm{z^k} = \frac{\sum_{i,j} \bm{\alpha^k}(i,j)\bm{M^k}(i,j)}{\sum_{i,j}\bm{\alpha^k}(i,j)}
\end{equation}
The retrieved feature set for class $k$ is now updated as $\mathcal{A}_k = \{\bm{z_i^k}\}_{i=1}^{N_a}$. We denote $\mathcal{A}=\cup_k\mathcal{A}_k$ as the set of augmented part features for all classes.

\paragraph{Refining Target Classifier.} 
With augmented part features, we can now refine the initial classifier with both support set $\mathcal{S}$ and augmented set $\mathcal{A}$ to get a more robust classifier. Since augmented features are not from the exactly same classes as support samples, we adopt the label smoothing technique \cite{Szegedy2016RethinkingTI} to prevent overfitting on the augmented features. Specifically, we convert an one-hot label $y$ into a smoothed probability distribution $\bm{p_y}$:
\begin{equation}
    \bm{p_y}(k) = 
    \begin{cases}
    1 - \epsilon \quad \quad \quad  ,k=y\\
    \frac{\epsilon}{|\mathcal{C}_{novel}|-1} \quad ,k\not= y
    \end{cases}
\end{equation}
where $\bm{p_y}(k)$ is the probability of the class $k$, and $\epsilon$ is a hyper-parameter in range $(0,1)$. Finally, the total loss for refining the classifier is:
\begin{equation}
 \mathcal{L}_{cls} = \sum_{\bm{z}\in\mathcal{S}} -\log p(y|\bm{z}) + \lambda \sum_{\bm{z}\in\mathcal{A}} KL(p(\cdot|\bm{z}), \bm{p_y})
\end{equation}
where the first term is cross entropy loss for samples in support set $\mathcal{S}$, and the second term is K-L divergence between predicted probability distribution $p(\cdot|\bm{z})$ and smoothed ground-truth distribution $\bm{p_y}$ for augmented features.

\section{Experiments}

\subsection{Datasets}
We conduct experiments on two widely-used few-shot learning datasets, namely \emph{mini}ImageNet and \emph{tiered}ImageNet. 
\paragraph{\emph{mini}ImageNet.} \emph{mini}ImageNet is a standard benchmark for few-shot learning proposed by \cite{vinyals2016matching}. It is a subset of the ImageNet \cite{ILSVRC15} and contains 100 classes and 600 examples for each class. We follow the protocol in \cite{Ravi2017OptimizationAA} to use 64 classes for training, 16 classes for validation and 20 classes for test. 

\paragraph{\emph{tiered}ImageNet.} \emph{tiered}ImageNet \cite{ren2018metalearning} is a larger subset of ImageNet and contains 608 classes and ~1000 images in each class. Theses classes are grouped into 34 higher categories, where 20 categories (351 classes) for training, 6 categories (97 classes) for validation and 8 categories (160 classes) for test. The large semantic difference between categories makes it more challenging for few-shot learning. 

\subsection{Implementation Details} 
For PDN, we transform input images with random crop, horizontal flip, color jitter and Gaussian blur. The crop scales for part view and global view are in range of (0.05, 0.14) and (0.14, 1), respectively. The number of cropped parts is set as $n=6$ by cross validation. The size of $\mathcal{D^-}$ is 1024 and 10240 for \emph{mini}ImageNet and \emph{tiered}ImageNet, respectively. We use standard ResNet-50 as backbone and resize images to $224\times224$. We set hyper-parameter $m=0.999$ and $\tau=0.2$. We adopt SGD optimizer with cosine learning rate decay. The learning rate is 0.015 for \emph{mini}ImageNet and 0.03 for \emph{tiered}ImageNet.

For PAN, we retrieve $N_a=1024$ extra images for each class. The label smoothing hyper-parameter $\epsilon$ is 0.2 for 1-shot and 0.7 for 5-shot. The loss weight $\lambda$ is set as 1. The classifier is trained with Adam and the learning rate is 0.001. Following the similar setting to \cite{vinyals2016matching}, We evaluate our method on 600 test episodes.

\begin{table}[t]
\centering
\resizebox{1\linewidth}{!}{
\begin{tabular}{ccccc}
\toprule
PDN w/o Select & PDN & PAN w/o $\rm C^2AM$ & PAN & Acc. \\
\midrule
 & & & & 57.12 \\
\checkmark & & & & 59.32 \\
\checkmark & & & \checkmark & 61.74 \\
 & \checkmark & & & 61.77 \\
 & \checkmark & \checkmark & & 62.64\\
 & \checkmark & & \checkmark & \textbf{63.84}\\
\bottomrule
\end{tabular}
}
\caption{Ablation study on \emph{mini}ImageNet in 5-way 1-shot.}
\label{tab:ablation}
\end{table}

\subsection{Comparison with Prior Work}
In Table \ref{tab:comparison}, we compare our method with both supervised and unsupervised few-shot learning methods. Overall, our method achieves the best performance under unsupervised setting, and is comparable with state-of-the-art supervised methods.

Our PDA-Net significantly outperforms unsupervised meta-learning \cite{hsu2018unsupervised} by 23.91\% and 29.14\% on \emph{mini}ImageNet in 1-shot and 5-shot, respectively. Compared with the most related method which also employs MoCo for representation learning \cite{10.1007/978-3-030-58568-6_16}, we achieve 9.65\% and 10.07\% improvement in 1-shot and 5-shot on \emph{miniImageNet}. On the larger tieredImageNet, our method largely outperforms a recent GAN-based method, GdBT2 \cite{Khoi2021SSLGAN}, by 21.15\% and 16.50\% in 1-shot and 5-shot, respectively. The significant improvements over the compared unsupervised methods verify the effectiveness of our PDA-Net.

Furthermore, our method outperforms two supervised meta-learning methods, LEO \cite{rusu2018metalearning} and MetaOptNet \cite{Lee2019MetaLearningWD}, on both datasets in both 1-shot and 5-shot. Compared with the supervised transfer learning method, Distill \cite{10.1007/978-3-030-58568-6_16}, the performance of our PDA-Net only drops 0.98\% on \emph{mini}ImageNet in 1-shot, which is acceptable considering the expensive label budget.
Alignment \cite{Afrasiyabi2020AssociativeAF} uses base labels to detect the most related class, which achieves better performance in 1-shot, but is comparable with ours in 5-shot.

\subsection{Further Analysis}
\paragraph{Ablation Study.} We evaluate the effectiveness of each key component of PDA-Net on \emph{mini}ImageNet in 1-shot, and show the results in Table \ref{tab:ablation}. The baseline is constructed by ablating part discovery and part augmentation, thus only taking as input two global views for contrastive learning. PDN w/o Select means that we use all the 6 parts for contrastive learning, while PAN w/o $\rm C^2AM$ means that we use class activation map instead \cite{zhou2016learning}.

Compared with the baseline, PDN w/o Select obtains 2.20\% performance gain, which verifies the advantage of part-based representation learning. PDN can get further improvement via selecting the most discriminative part. Better results are achieved by adding PAN to PDN or PDN w/o Select, which illustrates the effectiveness of our part augmentation strategy. Compared with CAM in \cite{zhou2016learning}, our $\rm C^2AM$ can obtain better performance, demonstrating the effectiveness of our proposed class competitive mechanism.



\begin{figure}
\begin{minipage}[b]{0.5\linewidth}
\centering
\includegraphics[width=0.92\linewidth]{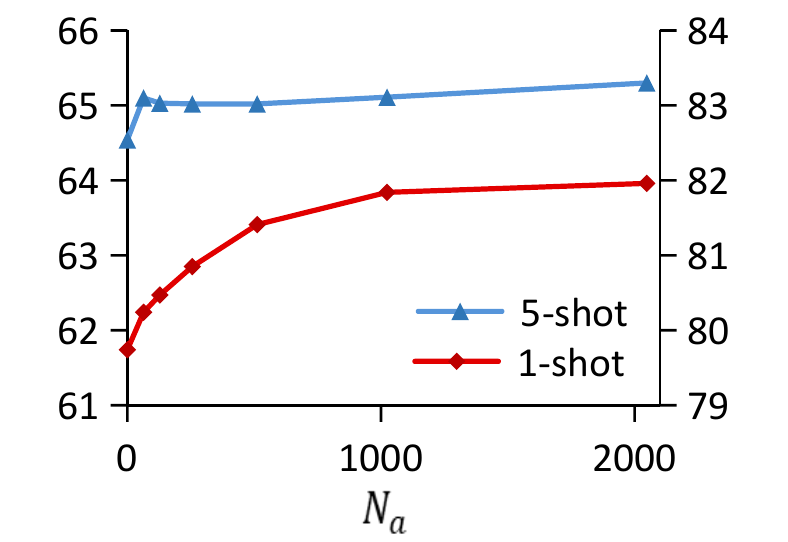}
\caption{Accuracy with different number of augmented features in PAN on $mini$ImageNet.}
\label{fig:Na}
\end{minipage}
\hspace{0.8mm}
\begin{minipage}[b]{0.45\linewidth}
\centering
\begin{tabular}{c|cc}
\toprule
$n$ & 1-shot & 5-shot \\
\midrule
2 & 57.84 & 76.38  \\
4 & 60.98 & 80.10 \\
6 & \textbf{63.84} & \textbf{83.11} \\
8 & 63.56  & 82.58  \\
\bottomrule
\end{tabular}
\tabcaption{Accuracy with different number of cropped parts in PDN on $mini$ImageNet.}
\label{tab:crop}
\end{minipage}
\end{figure}

\paragraph{Number of Crops.} In PDN, we use random crop to get $n$ parts. Here we evaluate the impact of number of cropped parts on representation learning. As shown in Table \ref{tab:crop}, the performance rapidly increases with the growing number of cropped parts, because it is more likely to contain discriminative parts with more cropped parts. The performance gets saturated after reaching 6, which indicates that our selection strategy is an effective way to extract discriminative parts.

\begin{figure}[t]
    \centering
    \includegraphics[width=6.5cm]{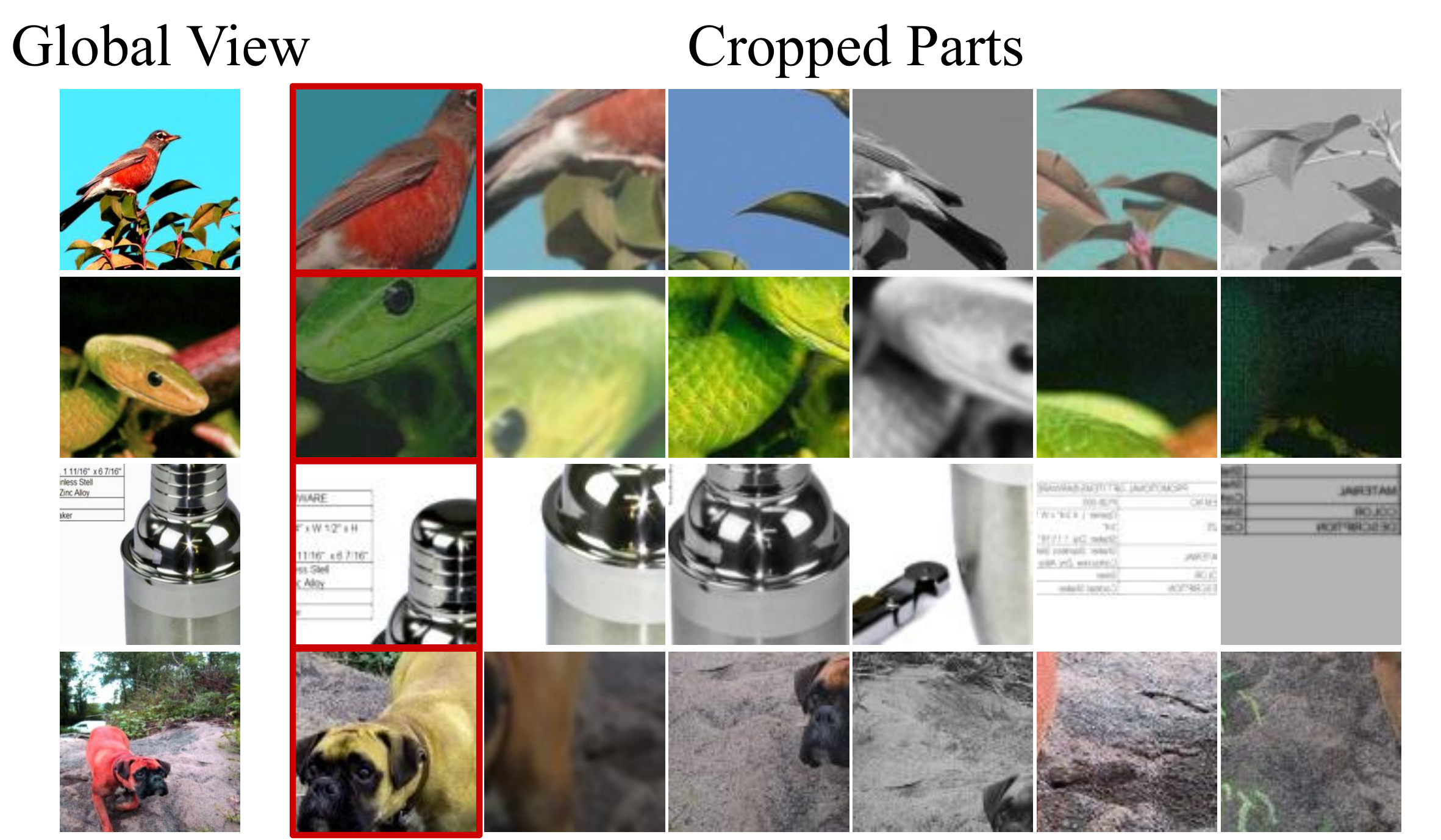}
    \caption{Visualization of the global view and cropped parts. }
    \label{fig:crops}
\end{figure}

\begin{figure}[t]
    \centering
    \includegraphics[width=5cm]{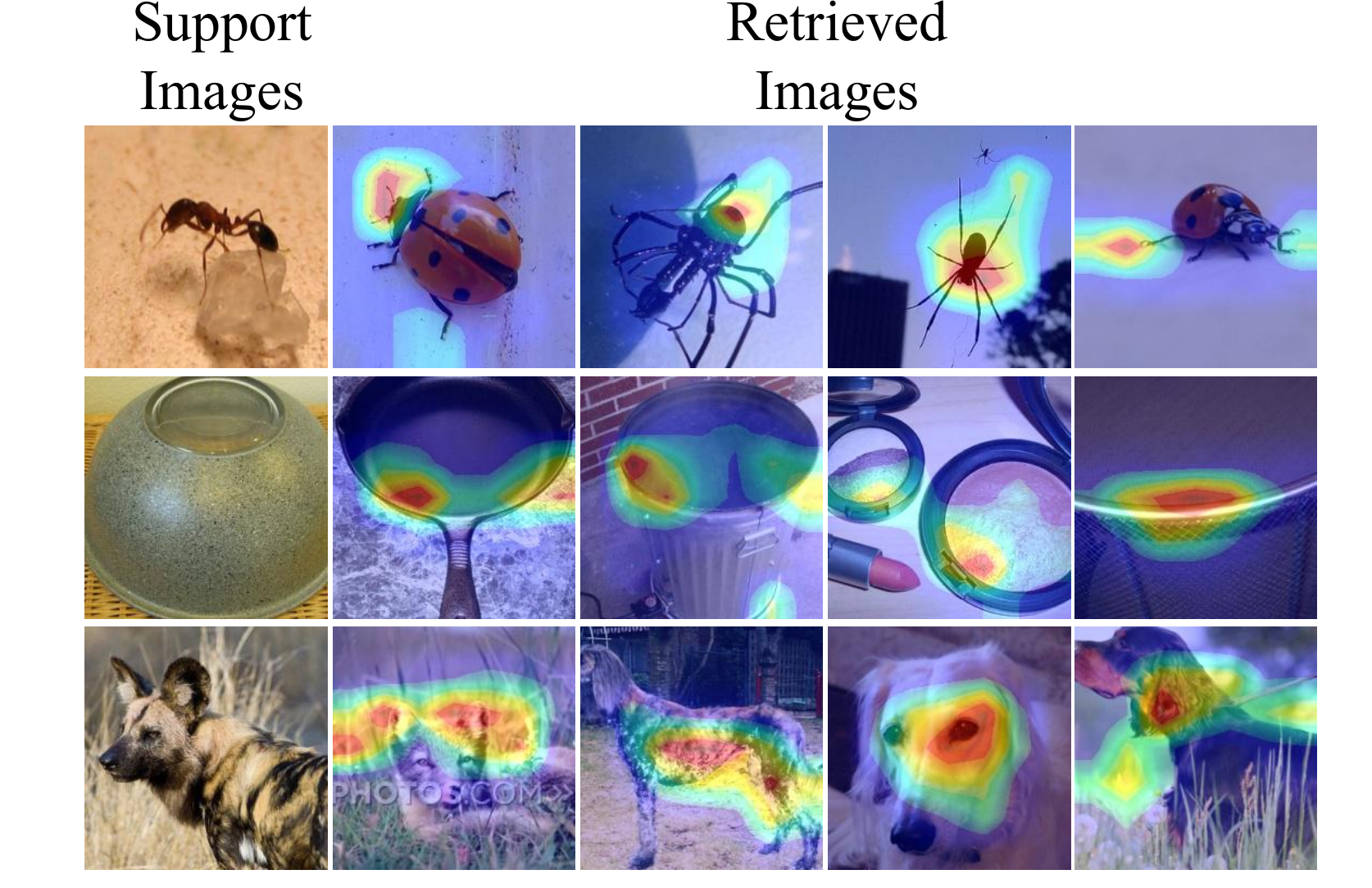}
    \caption{Visualization of retrieved images and their $\rm C^2AMs$.}
    \label{fig:attention}
\end{figure}

\paragraph{Number of Augmented Features.} In PAN, we retrieve $N_a$ extra images to augment each novel class. Here we experiment with different $N_a$ and show the results in Figure \ref{fig:Na}. It can be seen that the benefit of augmented features is more significant for 1-shot than 5-shot.
For 1-shot, the performance grows rapidly as $N_a$ increases and finally gets saturated after $N_a=1024$, which is because relevant features in base classes are limited. With a larger base dataset, we can infer that our method can get better results.

\subsection{Visualization}
To explore the discriminative part selection, we visualize the cropped parts and their corresponding global view in Figure \ref{fig:crops}. It should be noted that these images are all randomly transformed, so they may have different appearances even from the same input image.
We sort the cropped parts based on the distances to negative samples and select the parts with the largest distances within red boxes. For example, the main body of a bird and the head of a dog are selected in the first and fourth rows, respectively.
We can see that as the distances decrease, the parts are more likely to belong to background and contain less information about target object. 

To illustrate the rationality  of part augmentation strategy, we visualize the retrieved unlabeled images and their class-competitive attention maps in Figure \ref{fig:attention}. We can see that the retrieved images are usually similar to support images. More interestingly, $\rm C^2AM$ can locate class-specific image regions which are very relevant to the objects in support set. For example, the antennae and legs of the ladybug are highlighted in the retrieved image, which are very similar to those of the ant in support image. In contrast, its shell attracts less attention due to the distinctive appearance from the ant.

\section{Conclusion}
In this paper, we present a Part Discovery Network, which can extract effective prior knowledge for few-shot learning from a flat collection of unlabeled images. Furthermore, a Part Augmentation Network is proposed to augment support examples with relevant parts, which can mitigate overfitting and lead to better classification boundaries. The experimental results demonstrate that our method significantly outperforms previous unsupervised meta-learning methods and achieves comparable accuracy with state-of-the-art supervised methods.

\section*{Acknowledgements}
This work is supported by National Natural Science Foundation of China (61976214, 61633021, 61721004
, 61836008), Beijing Municipal Natural Science Foundation (4214075), and Science and Technology Project of SGCC Research on feature recognition and prediction of typical ice and wind disaster for transmission lines based on small sample machine learning method.


\bibliographystyle{named}
\bibliography{main}

\end{document}